\documentclass{article}
\usepackage{spconf,amsmath,graphicx}

\usepackage{algorithm}
\usepackage{algpseudocode}
\algnewcommand{\LineComment}[1]{\State \(\triangleright\) #1}
\usepackage[colorinlistoftodos,prependcaption,textsize=tiny]{todonotes}

\title{Learning to integrate vision data into road network data}
\name{Oliver Stromann$^{\star \dagger}$ \qquad Alireza Razavi$^{\dagger}$ \qquad Michael Felsberg$^{\star}$\thanks{We gratefully acknowledge the support of Sweden’s innovation agency, Vinnova, through project iQDeep (project number 2018-02700). Computations were enabled by resources provided by the SNIC at C3SE partially funded by the Swedish Research Council through grant agreement no. 2018-05973.}}

\address{$^{\star}$ Computer Vision Laboratory, Linköping University, Sweden \\
    $^{\dagger}$Scania CV AB, Autonomous Transport Solution Research, Sweden }
    
\begin{document}
%
\maketitle
\begin{abstract}
Road networks are the core infrastructure for connected and autonomous vehicles, but creating meaningful representations for machine learning applications is a challenging task. In this work, we propose to integrate remote sensing vision data into road network data for improved embeddings with graph neural networks. We present a segmentation of road edges based on spatio-temporal road and traffic characteristics, which allows enriching the attribute set of road networks with visual features of satellite imagery and digital surface models. We show that both, the segmentation and the integration of vision data can increase performance on a road type classification task, and we achieve state-of-the-art performance on the OSM+DiDi Chuxing dataset on Chengdu, China.
\end{abstract}
\vspace{-2mm}
\section{Introduction}
\vspace{-2mm}

Knowledge about road networks (RNs) and the traffic flowing through it is key for making good decisions for connected and autonomous vehicles. It is therefore of large importance to find meaningful and efficient representations of RNs to allow and ease the knowledge generation.
 
However, the structural information encoded in spatial network data like RNs has two shortcomings: first, the incompleteness of the encoded information and second the absence of information that is difficult to encode yet still relevant. On the other hand, there exist vast amounts of unstructured image data that contains complementary data. Therefore, we propose to enrich the structured but incomplete spatial network data with unstructured but spatially complete data in the form of continuous imagery. We demonstrate the integration of vision data into a Graph Neural Network (GNNs) by low-level visual features and propose a segmentation of the road network graph according to spatial and empirical traffic constraints.

Specifically, we address the challenge of enriching crowd-sourced RN data from OpenStreetMap (OSM)~\cite{OpenStreetMap} with remotely sensed 
data from Maxar Technologies~\cite{maxar2021}. We do so by utilizing GPS tracks from DiDi Chuxing's ride hailing service~\cite{didigaia2021} to spatially segment the RN based on empirical travel times. We evaluate our proposed method on a node classification task for road type labels using GraphSAGE~\cite{hamilton2017inductive}.

The results confirm that our approach leads to improved performance on a classification of road type labels in both supervised and unsupervised learning settings.
To summarize, our contributions are: 1) integrating image data through low-level visual features into a graph representation of spatial structures by means of spatial segmentation 2) a systematic evaluation of our proposed method in a supervised and unsupervised learning of node classifications.


\vspace{-2mm}
\section{Related Work}
\vspace{-2mm}


\subsection{Geographical Data on Road Networks}
\vspace{-2mm}

Graph data on RNs is spatial network data, which is nowadays easily accessed through common mapping sources. 
The RN is represented as an attributed directed spatial graph $G{=}(V,E)$, with intersections as nodes $v \in V$ and roads connecting these intersections as edges $e \in E$.
From the crowd-sourced OSM~\cite{OpenStreetMap}, intersection attributes 
, and road attributes 
can be obtained. 
One shortcoming of the available crowd-sourced RN data is that some attributes are not consistently recorded. Either data is missing completely for certain geographical regions, or it is inconsistently set~\cite{funke2015automatic}. 

Remote sensing data is data collected from air- or space borne sensors like radars, LiDARs or cameras. It requires extensive data preprocessing like atmospheric, radiometric and topographic error correction. Nowadays, analysis-ready remote sensing data is available which has undergone correction. 

\vspace{-4mm}
 \subsection{Machine Learning Concepts}
 \vspace{-2mm}
 \subsubsection{Graph Neural Networks}
 \vspace{-2mm}
 GNNs experienced a surge in recent years. Many architectures have been proposed to produce deep embeddings of nodes, edges, or whole graphs~\cite{hamilton2020graph-rep,zhou2020graph}. Techniques from other deep learning domains such as computer vision and natural language processing have been successfully integrated into GNN architectures~\cite{zhou2020graph}. 
 In our work we focus on learning node embeddings with GraphSAGE~\cite{hamilton2017inductive} - a GNN architecture which relies on an efficient sampling of neighbour nodes and an aggregation of their features.
 \vspace{-4mm}
 \subsubsection{Visual Feature Extraction}
 \vspace{-2mm}
 Visual feature extraction is the process of obtaining information from images to derive informative, non-redundant features to facilitate subsequent machine learning tasks. Convolutional Neural Networks (CNNs) replaced hand-craft extractors in most applications. CNNs either require data of the application to train or need to be pre-trained on enormous datasets to then be transferred to the application domain~\cite{pires2020cnnTransferRS}.
 We explicitly evaluate a simple pixel statistic - intensity histograms~\cite{gonzalez2002digital} - in our approach.
 
 

 \vspace{-4mm}
 \subsection{Machine Learning on Geographical Data}
 \vspace{-1mm}

 \subsubsection{Machine Learning on Road Networks}
 \vspace{-2mm}
Examples of machine learning tasks that are applied to RNs range method-wise from classification and regression to sequence-prediction and clustering~\cite{gharaee2021graph, jepsen2020relational, wu2020hrnr}, and application-wise from vehicle-centric predictions such as next-turn, destination and time of arrival predictions or routing to RN-centric predictions such as speed limit, travel time or traffic flow predictions~\cite{wu2020hrnr,liu2020graphsage}. 

Following the growing popularity of OSM, several machine learning methods have been proposed in the recent years to either improve or to use OSM data~\cite{vargas2020openstreetmap}. For RNs, conventional machine learning techniques that do not exploit the graph structure have been used to predict road type labels or to impute missing attributes and topologies~\cite{funke2015automatic,Jasmeet2018quality_osm}. 

GNNs offer the advantage that graph topologies are directly exploited, and no additional features need to be constructed. Consequently, several authors have demonstrated the effectiveness of GNNs on RNs in classifications~\cite{jepsen2020relational, wu2020hrnr, he2020roadtagger}, regressions~\cite{jepsen2020relational} and sequence predictions~\cite{wu2020hrnr}. Jepsen \textit{et al.}~\cite{jepsen2020relational} propose a relational fusion network (RFN), which use different representations of a road network concurrently to aggregate features. Wu \textit{et al.}~\cite{wu2020hrnr} developed a hierarchical road network representation (HRNR) in which a three-level neural architecture is constructed that encodes functional zones, structural regions and roads respectively.
He \textit{et al.}~\cite{he2020roadtagger} proposed an integration of visual features from satellite imagery through CNNs as node features to a GNN.
\vspace{-1mm}
\subsubsection{Visual Feature Extraction in Remote Sensing}
\vspace{-2mm}
As in many other domains, transfer learning has also been applied in remote sensing to address common tasks like scene classification~\cite{pires2020cnnTransferRS} or object detection~\cite{chen2018transferRS}. 
On the other hand, hand-crafted features combined with classical machine learning algorithms are still commonly used in computer vision on remote sensing. Object-based land cover classification frequently uses hand-crafted texture or intensity histogram features and still achieve state-of-the-art performance~\cite{stromann2020dimensionality, tassi2020object, verde2020national}. 

\vspace{-2mm}
\section{Materials \& Methods}
\vspace{-2mm}

\subsection{Road Network Graph}
\vspace{-2mm}
Our datasets consist of RN data, GPS tracks and remote sensing imagery.
We follow the commonly used approach of dual graph representation $G^D{=}(E, B)$ of RNs~\cite{jepsen2020relational, gharaee2021graph, wu2020hrnr, liu2020graphsage}. That means, the graph consists of road segments as nodes $e \in E$ and connection between roads segments as edges $b \in B$.

In this study, RN data is obtained from OSM~\cite{OpenStreetMap} using OSMnx~\cite{boeing2017osmnx}. 
We add traffic information to the road attributes by matching GPS tracks of ride hailing vehicles to the RN~\cite{yang2018fmm}. 


\vspace{-4mm}
\subsection{GraphSAGE}
\vspace{-2mm}
GraphSAGE~\cite{hamilton2017inductive} is a graph convolutional neural network, which for an input graph $G{=}(V,E)$ produces node representations $\mathbf{H}_v^{k}$ at layer depth $k$. This is done by aggregating features from neighbouring nodes $N(v)$ using an aggregation function $\textsc{Agg}^k$ and after a linear matrix multiplication with a weight matrix $\mathbf{W}^k$ apply a non-linear activation function $\sigma$, such that
\begin{equation}
    \label{eq:aggregation}
    \mathbf{H}_v^{k} = \sigma\left(\mathbf{W}^k \cdot
        \textsc{Agg}^k\left(\{\mathbf{H}_n^{(k-1)}\vert n\in N(v)\}\right)
    \right).
\end{equation}

At layer depth $k{=}0$ the aggregated features consist of the neighbouring nodes' attributes. The node representations $\mathbf{H}_v^{K}$ after the last layer (i.e., $k{=}K$) can be used as node embeddings $\mathbf{z}_v$ which serve as input for a downstream machine learning task, like a classifier in our case. The interested reader is referred to Hamilton \textit{et al.}~\cite{hamilton2017inductive} for implementation details of GraphSAGE.

\begin{figure}[]
    \centering
    \begin{tabular}{c c c }
         \includegraphics[angle=90, origin=c,width=0.28\linewidth]{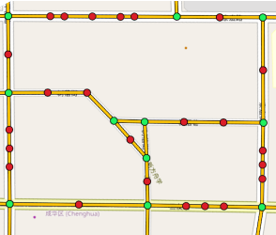} & \includegraphics[angle=90, origin=c,width=0.28\linewidth]{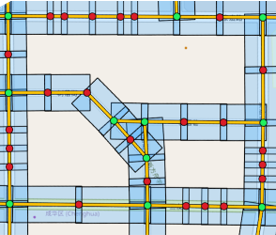} &
         \includegraphics[angle=90, origin=c,width=0.28\linewidth]{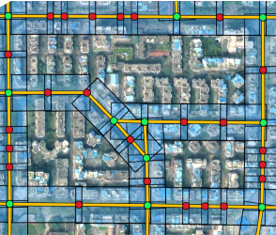} \\
         a) & b) &  c) \\
    \end{tabular}
    \caption{Processing steps of our proposed method. a) Graph $G_{ORN}$ with intersections $V_{ORN}$ in green and roads $E_{ORN}$ in yellow. Overlaid is $G_{SRN}$ with interstitial vertices in red. b) Rectangular road segment footprints in blue. c) \textit{TrueOrtho} RGB-imagery. Pixels within each blue rectangle are added to the road segment attributes.}
    \label{fig:workflow}
\end{figure}

\vspace{-4mm}
\section{Contributions}
\vspace{-2mm}
We propose to enrich the attribute set of spatial network data of RNs with visual features from remote sensing data and present a segmentation of the RN into a fine-grained graph representation. Figure~\ref{fig:workflow} outlines the processing steps of our proposed method. 

\vspace{-1mm}
\subsection{Segmentation}
\label{section:segmentation}
\vspace{-2mm}
While the topological representation of an RN is advantageous for routing applications, we argue that a more fine-grained representation of an RN is better suited for many machine learning tasks. In a topological RN, road geometries may vary in length, and road attributes might be non-static for the entirety of a road. 
Therefore, we introduce a segmentation which creates a more fine-grained representation of the RN with shorter edges. To take the spatio-temporal nature of traffic on the RN into account, the segmentation aims to create segments of target travel times as well as target segment lengths. 

\begin{algorithm}[tbh]
    \caption{Segmentation Algorithm.}
    \label{alg:segment}
    \begin{algorithmic}[1]
        \Procedure{Segment}{$G = (V, E), \mathit{attr}, \mathit{target_{attr}}$} \\
        \Comment{Input graph $G$, edge attribute $\mathit{attr}$ and target value $\mathit{target_{attr}}$}
            \For{$(u,v,a) \in E$}
                \State $n\gets \lceil a_\mathit{attr}/\mathit{target_{attr}} \rceil$\\  \Comment{Calculate number of segmentations $n$}
                \State $u_1 \gets u$
                \For{$i\gets 1, n$}
                    \State $v_i \gets $\Call{interpolate}{$a_{\mathit{geom}}, \frac{i}{n}$}\\ \Comment{Add node $v_i$ at $\frac{i}{n}$-distance of $a_{\mathit{geom}}$ between $u$ and $v$}
                    \State $V \gets V \cup v_i$
                    \State $E \gets E \cup (u_i,v_i)$ 
                    \State $u_{i+1} \gets v_i$ 
                \EndFor
            \State $E \gets E \backslash (u,v)$ 
            \State $V \gets V \backslash (u,v)$ 
            \EndFor
            \State
            \textbf{return} $G$
        \EndProcedure
    \end{algorithmic}
\end{algorithm}

Given a target segment travel time $\mathit{target_{traveltime}}$ and a target segment length $\mathit{target_{length}}$, we determine the number of equally distanced split points along a road geometry. At these points, interstitial nodes are inserted, replacing original edges with shorter edges. From the original RN $G_{\mathit{ORN}}$, the segmented RN $G_{\mathit{SRN}}$ is thus created by running Algorithm~\ref{alg:segment} first with $\mathit{attr}{=}\mathit{traveltime}$ and secondly with $\mathit{attr}{=}\mathit{length}$.
The segmentation is done prior to the conversion of graph $G$ to its dual representation $G^D$. 
\vspace{-1mm}
\subsection{Integration of Vision Data}
\vspace{-2mm}
Though the traffic information and segmentation enriches the attribute sets, the structural information in the spatial network of $G$ still suffers from incompleteness in the encoded information, as not all attributes are consistently set in the crowd-sourced data from OSM. Furthermore, potentially relevant information in the vicinity of the spatial network are not covered in the attribute sets. 

Continuous image data on the other hand has the potential to capture such information. Hence, we enrich the road attribute set with visual features from remote sensing data. Analysis-ready high-resolution orthorectified satellite imagery and a Digital Surface Model (DSM) are used in our study.
We use a rectangle around a road footprint (Figure\ref{fig:workflow} b)) to extract image patches of the RGB imagery and the DSM (Figure\ref{fig:workflow} c)). From each image patch, we use intensity histograms per channel as visual features. The frequencies in each histogram bin are used as numerical features in the road attributes. These features have the benefit compared to CNNs that they require no additional training.


\vspace{-1mm}
\subsection{Summary}
\vspace{-2mm}

We present a method to enrich structural, but incomplete, spatial network data with less-structured, but spatially complete data in the form of continuous imagery. We rely on simple low-level visual features, that require no learning, and demonstrate the effectiveness of this approach in the context of crowd-sourced RN data and high-resolution remote sensing imagery. Our work is closely related to RoadTagger by He \textit{et al.}~\cite{he2020roadtagger}. However, RoadTagger relies purely on CNNs to extract visual features and trains them end-to-end with the GNN, whereas we fuse low-level visual features with the road network attributes from OSM. Moreover, we extend the experiments also to unsupervised learning and demonstrate that it can achieve comparable performance on the binary road type classification problem. In general, our proposed method offers a light-weight alternative to CNN-based approaches, while achieving comparable performances.

\vspace{-2mm}
\section{Experiments}
\vspace{-2mm}

\subsection{Datasets}
\vspace{-2mm}
An RN from the city of Chengdu, China is extracted from OSM~\cite{OpenStreetMap}. We match GPS trajectories from ride-hailing vehicles 
to the RN. 
We use high-resolution (0.5 m/pixel) analysis-ready orthorectified satellite imagery (\textit{TrueOrtho}) and a Digital Surface Model (\textit{DSM}) as vision data. 

We construct three datasets: The unsegmented, original RN (\textsc{ORN}), the segmented RN (\textsc{SRN}) and the segmented RN visual features (\textsc{SRN+Vis}).  
In \textsc{SRN} and \textsc{SRN+Vis}, interstitial nodes are inserted according to the proposed segmentation in Algorithm~\ref{alg:segment}. We set the target travel time $target_{traveltime}$ to 15 s and target segment length $target_{length}$ to 120 m. 
In \textsc{ORN}, we randomly sample 20\% of the nodes in $E$ for validation and 20\% for testing. The remaining nodes are used for training. Validation and test set allocations are propagated down from \textsc{ORN} to \textsc{SRN} and \textsc{SRN+Vis}. 


The following features make up the feature set:
Geographical features of \textit{length}, \textit{bearing} \textit{centroid}, \textit{geometry} which is the road geometry resampled to a fixed number of equally-distanced points and translated by the \textit{centroid} to yield relative distances in northing and easting (meters). Binary features depicting \textit{one-way}, \textit{bridge} and \textit{tunnel}. 
GPS features from the matched trajectories, consisting of \textit{travel times} as median travel times and \textit{throughput} as average daily vehicle throughput. 
Additionally, we extract visual features from image patches of 120 m by 120 m from \textit{TrueOrtho} and \textit{DSM} following the \textit{bearing}. 
In \textsc{SRN+Vis} the pixel values of each channel (3 RGB channel for the \textit{TrueOrtho}, 1 graylevel channel for \textit{DSM}) are binned into a histogram of 32 bins.

\vspace{-4mm}
\subsection{Training}
\vspace{-2mm}

To demonstrate how the RN segmentation and the integration of visual features improve the node embedding, we train for node classifications of road type labels.

The \textit{highway} label from OSM is used as the target. The label describes the type of road as an indicator of road priority. 
The classes are \textit{motorway}, \textit{trunk}, \textit{primary}, \textit{secondary}, \textit{tertiary}, \textit{unclassified}, \textit{residential} and \textit{living street}. \textit{Living street} and \textit{motorway} are underrepresented in both ORN and SRN with  less than 2\% of all samples. This class imbalance makes the 8-class classification a challenging problem.
Additionally, to set this work into context of RoadTagger~\cite{he2020roadtagger}, we perform a binary classification, by aggregating the 8-class predictions of the first four classes (\textit{motorway} to \textit{secondary}) and the remaining four classes (\textit{tertiary} to \textit{living street}) to two classes.



We train a GraphSAGE model with layer depth $K=2$ and mean pooling aggregators using Adam optimizer.
Model selection is based on validation performance and reported are test performances. 
We perform a Bayesian hyperparameter search~\cite{wandb}. 
The search space of hyperparameters is composed of \textit{hidden units} $\in \{512, 1024\}$, \textit{embedding dimensionality} $\in \{2^3, 2^4, ..., 2^7\}$,  \textit{learning rate} $\in [1e-8, 1e-1]$, \textit{weight decay} $\in [0, 0.1]$ and \textit{dropout rate} $\in [0, 0.4]$.
Unsupervised models are trained for 20 epochs with a batch size of 1024. Supervised settings are trained for 100 epochs with a  batch size of 512.

\vspace{-2mm}
\section{Numerical Results}
\vspace{-2mm}

Tables~\ref{tab:results_supervised} and~\ref{tab:results_unsupervised} show the test results on road type classification in micro-averaged F1-Scores for supervised and unsupervised learning respectively. 
A majority voting from \textsc{SRN} and \textsc{SRN+Vis} to \textsc{ORN} is stated. The first two rows depict performance on the 8-class classification problem of eight road type labels. The performance on the binary classification is depicted in the last two rows.

The supervised results in table~\ref{tab:results_supervised}, the majority vote of the different \textsc{SRN} subsets shows clearly that \textsc{SRN+Vis} produces the best performing model with a 13.5\% improvement compared to \textsc{ORN}. The segmentation alone (\textsc{SRN}) improves the performance only to a small extent of 1.4\%. When aggregating the predictions to two classes, \textsc{SRN+Vis} again achieves the best performance with a 0.7\% improvement over \textsc{ORN}. 

In the unsupervised setting in table~\ref{tab:results_unsupervised}, 
the majority voting from \textsc{SRN} is on par with \textsc{ORN}, while \textsc{SRN+Vis} achieves a small improvement of 0.4\%. For the binary classification, \textsc{SRN} shows a small performance decrease of 0.7\%, while \textsc{SRN+Vis} improves the classification by 0.2\%. Curiously, the unsupervised node classification on the binary classification exceeds performance of supervised node classification.  

Overall, the performances on binary classification of 0.911 and 0.915 in supervised and unsupervised respectively, are comparable with the performance of RoadTagger~\cite{he2020roadtagger}, which achieved an accuracy of 93.1\% on a similar dataset.

\begin{table}[]
\centering
\caption{Results on supervised node classification.}\label{tab:results_supervised}
\begin{tabular}{|r|r|r|r|}
\hline
&  \textbf{\textsc{ORN}} & \textbf{\textsc{SRN}} & \vtop{\hbox{\strut \textbf{\textsc{SRN}}}\hbox{\strut \textbf{\textsc{+Vis}}}}\\ \hline
\hline
\multicolumn{1}{|r|}{\begin{tabular}[r]{@{}r@{}}GNN\\ + Majority Vote\end{tabular}} & 0.580 & 0.588 & \textbf{0.658} \\ 
\hline
\multicolumn{1}{|r|}{\begin{tabular}[r]{@{}r@{}}Percentage Gain\\ over ORN\end{tabular}} & - & 1.4\% & \textbf{13.5\%} \\ 
\hline 
\hline
\multicolumn{1}{|r|}{\begin{tabular}[r]{@{}r@{}}GNN (2-class) \\ + Majority Vote\end{tabular}}& 0.905 & 0.905 & \textbf{0.911} \\
\hline
\multicolumn{1}{|r|}{\begin{tabular}[r]{@{}r@{}}Percentage Gain\\ over ORN (2-class) \end{tabular}} & - & 0.0\% & \textbf{0.7\%} \\ 
\hline
\end{tabular}
\end{table}

\begin{table}[]
\centering
\caption{Results on unsupervised node classification.}\label{tab:results_unsupervised}
\begin{tabular}{|r|r|r|r|}
\hline
&  \textbf{\textsc{ORN}} & \textbf{\textsc{SRN}} & \vtop{\hbox{\strut \textbf{\textsc{SRN}}}\hbox{\strut \textbf{\textsc{+Vis}}}}\\ \hline
\hline
\multicolumn{1}{|r|}{\begin{tabular}[r]{@{}r@{}}GNN\\ + Majority Vote\end{tabular}} &  0.532 & 0.532  & \textbf{0.534} \\ 
\hline
\multicolumn{1}{|r|}{\begin{tabular}[r]{@{}r@{}}Percentage Gain\\ over ORN\end{tabular}}& - &  0.0\% &  \textbf{0.4\%}\\ 
\hline
\hline
\multicolumn{1}{|r|}{\begin{tabular}[r]{@{}r@{}}GNN (2-class) \\ + Majority Vote\end{tabular}} & 0.913 & 0.907 & \textbf{0.915} \\
\hline
\multicolumn{1}{|r|}{\begin{tabular}[r]{@{}r@{}}Percentage Gain\\ over ORN (2-class) \end{tabular}} & - & -0.7\% & \textbf{0.2\%} \\ 
\hline
\end{tabular}
\end{table}


\vspace{-2mm}
\section{Discussion \& Conclusion}
\vspace{-2mm}
We have presented a method to enrich incomplete, structural spatial network data with complete, continuous image data. In the example of road type classification on crowd-sourced RNs, we demonstrated how low-level visual features from remote sensing data can be included into the attribute set of the RN. Moreover, we presented a segmentation based on spatial and empirical traffic information.

The results of our experiments show, that low-level visual features like pixel intensity histograms, can improve performance on both supervised and unsupervised node classification of road type labels. Moreover, we have shown that \textsc{SRN+Vis} can achieve performances similar to state-of-the-art models for binary road type classifications in both supervised and unsupervised settings.

Future work, should investigate to what extent the replacement of low-level visual features through CNNs like in RoadTagger~\cite{he2020roadtagger} is beneficial for supervised and unsupervised node classifications in RN. Another interesting direction of research is to analyse how the unsupervised embeddings can be used for multiple different machine learning tasks that are relevant from a RN perspective.


\bibliographystyle{IEEEbib}
\bibliography{strings,refs}

\begin{thebibliography}{10}

\bibitem{OpenStreetMap}
{OpenStreetMap},
\newblock ``{Planet dump retrieved from https://planet.osm.org on
  2021-06-10},'' 2021 [Online].

\bibitem{maxar2021}
Maxar {Technologies},
\newblock ``https://www.maxar.com/,'' Aug. 2021 [Online].

\bibitem{didigaia2021}
The~{Gaia} {Initiative},
\newblock ``https://outreach.didichuxing.com/ appen-vue/,'' May 2021 [Online].

\bibitem{hamilton2017inductive}
William~L Hamilton, Rex Ying, and Jure Leskovec,
\newblock ``Inductive representation learning on large graphs,''
\newblock in {\em Proceedings of the 31st NeurIPS}, 2017, pp. 1025--1035.

\bibitem{funke2015automatic}
Stefan Funke, Robin Schirrmeister, and Sabine Storandt,
\newblock ``Automatic extrapolation of missing road network data in
  openstreetmap,''
\newblock in {\em Proceedings of the 2nd International Conference on Mining
  Urban Data-Volume 1392}, 2015, pp. 27--35.

\bibitem{hamilton2020graph-rep}
William~L. Hamilton,
\newblock ``Graph representation learning,''
\newblock {\em Synthesis Lectures on Artificial Intelligence and Machine
  Learning}, vol. 14, no. 3, pp. 1--159, 2020.

\bibitem{zhou2020graph}
Jie Zhou, Ganqu Cui, Shengding Hu, Zhengyan Zhang, Cheng Yang, Zhiyuan Liu,
  Lifeng Wang, Changcheng Li, and Maosong Sun,
\newblock ``Graph neural networks: A review of methods and applications,''
\newblock {\em AI Open}, vol. 1, pp. 57--81, 2020.

\bibitem{pires2020cnnTransferRS}
Rafael Pires~de Lima and Kurt Marfurt,
\newblock ``Convolutional neural network for remote-sensing scene
  classification: Transfer learning analysis,''
\newblock {\em Remote Sensing}, vol. 12, no. 1, pp. 86, 2020.

\bibitem{gonzalez2002digital}
Rafael~C Gonzalez, Richard~E Woods, et~al.,
\newblock {\em Digital image processing},
\newblock Prentice hall Upper Saddle River, NJ, 2002.

\bibitem{gharaee2021graph}
Zahra Gharaee, Shreyas Kowshik, Oliver Stromann, and Michael Felsberg,
\newblock ``Graph representation learning for road type classification,''
\newblock {\em Pattern Recognition}, p. 108174, 2021.

\bibitem{jepsen2020relational}
Tobias~Skovgaard Jepsen, Christian~S Jensen, and Thomas~Dyhre Nielsen,
\newblock ``Relational fusion networks: Graph convolutional networks for road
  networks,''
\newblock {\em IEEE Transactions on Intelligent Transportation Systems}, 2020.

\bibitem{wu2020hrnr}
Ning Wu, Xin~Wayne Zhao, Jingyuan Wang, and Dayan Pan,
\newblock ``Learning effective road network representation with hierarchical
  graph neural networks,''
\newblock in {\em Proceedings of the 26th ACM SIGKDD}, 2020, pp. 6--14.

\bibitem{liu2020graphsage}
Jielun Liu, Ghim~Ping Ong, and Xiqun Chen,
\newblock ``Graphsage-based traffic speed forecasting for segment network with
  sparse data,''
\newblock {\em IEEE Transactions on Intelligent Transportation Systems}, 2020.

\bibitem{vargas2020openstreetmap}
John~E Vargas-Munoz, Shivangi Srivastava, Devis Tuia, and Alexandre~X Falcao,
\newblock ``Openstreetmap: Challenges and opportunities in machine learning and
  remote sensing,''
\newblock {\em IEEE Geoscience and Remote Sensing Magazine}, vol. 9, no. 1, pp.
  184--199, 2020.

\bibitem{Jasmeet2018quality_osm}
Jasmeet Kaur and Jaiteg Singh,
\newblock ``An automated approach for quality assessment of openstreetmap
  data,''
\newblock in {\em 2018 International Conference on Computing, Power and
  Communication Technologies (GUCON)}, 2018, pp. 707--712.

\bibitem{he2020roadtagger}
Songtao He, Favyen Bastani, Satvat Jagwani, Edward Park, Sofiane Abbar,
  Mohammad Alizadeh, Hari Balakrishnan, Sanjay Chawla, Samuel Madden, and
  Mohammad~Amin Sadeghi,
\newblock ``Roadtagger: Robust road attribute inference with graph neural
  networks,''
\newblock in {\em Proceedings of the AAAI}, 2020, vol.~34, pp. 10965--10972.

\bibitem{chen2018transferRS}
Zhong Chen, Ting Zhang, and Chao Ouyang,
\newblock ``End-to-end airplane detection using transfer learning in remote
  sensing images,''
\newblock {\em Remote Sensing}, vol. 10, no. 1, pp. 139, 2018.

\bibitem{stromann2020dimensionality}
Oliver Stromann, Andrea Nascetti, Osama Yousif, and Yifang Ban,
\newblock ``Dimensionality reduction and feature selection for object-based
  land cover classification based on sentinel-1 and sentinel-2 time series
  using google earth engine,''
\newblock {\em Remote Sensing}, vol. 12, no. 1, pp. 76, 2020.

\bibitem{tassi2020object}
Andrea Tassi and Marco Vizzari,
\newblock ``Object-oriented lulc classification in google earth engine
  combining snic, glcm, and machine learning algorithms,''
\newblock {\em Remote Sensing}, vol. 12, no. 22, pp. 3776, 2020.

\bibitem{verde2020national}
Natalia Verde, Ioannis~P Kokkoris, Charalampos Georgiadis, Dimitris Kaimaris,
  Panayotis Dimopoulos, Ioannis Mitsopoulos, and Giorgos Mallinis,
\newblock ``National scale land cover classification for ecosystem services
  mapping and assessment, using multitemporal copernicus eo data and google
  earth engine,''
\newblock {\em Remote Sensing}, vol. 12, no. 20, pp. 3303, 2020.

\bibitem{boeing2017osmnx}
Geoff Boeing,
\newblock ``Osmnx: New methods for acquiring, constructing, analyzing, and
  visualizing complex street networks,''
\newblock {\em Computers, Environment and Urban Systems}, vol. 65, pp.
  126--139, 2017.

\bibitem{yang2018fmm}
Can Yang and Gyozo Gidofalvi,
\newblock ``Fast map matching, an algorithm integrating hidden markov model
  with precomputation,''
\newblock {\em International Journal of Geographical Information Science}, vol.
  32, no. 3, pp. 547--570, 2018.

\bibitem{wandb}
Lukas Biewald,
\newblock ``Experiment tracking with weights and biases,'' Aug. 2021 [Online],
\newblock Software available from wandb.com.

\end{thebibliography}

\end{document}